\documentclass[conference]{IEEEtran}
\usepackage{amsmath}
\usepackage{array}
\usepackage{easylist}
\usepackage{graphicx}       
\graphicspath{{imgs/}}
\usepackage{mathtools}
\usepackage{placeins}
\usepackage{xcolor}
\usepackage{float}

\usepackage[dvipsnames]{xcolor}
\usepackage{soul}
\usepackage[normalem]{ulem}

\usepackage[utf8]{inputenc}
\usepackage{ifthen}
\usepackage{xurl}
\usepackage{hyperref}

\usepackage[dvipsnames]{xcolor}
\usepackage{soul}
\usepackage[normalem]{ulem}

\usepackage{epsf,picinpar}
\usepackage{varioref}
\usepackage{fdsymbol}

\usepackage[backend=biber,style=ieee,maxcitenames=2,maxbibnames=99,citestyle=numeric-comp]{biblatex}

\usepackage{pifont}

\usepackage{orcidlink}
\usepackage{booktabs}

\usepackage{listings}

\usepackage{multicol}
\usepackage{multirow}
\usepackage{makecell}
\usepackage{caption}
\usepackage{subcaption}
\usepackage[framemethod=TikZ]{mdframed}

\usepackage{dblfloatfix}
\usepackage{nicefrac}
\usepackage{framed}

\usepackage{fontawesome5}

\usepackage{makecell}

\usepackage{url}

\usepackage{longtable}
\usepackage{afterpage}

\newcommand{\assignedto}[1]{%
    \ifthenelse{\boolean{showannotations}}%
    {\textbf{\noindent\ding{46}\textcolor{white}{~\colorbox{\assignementcolor}{Assigned to:}}~\textcolor{\assignementcolor}{#1}\\}%
    }
    {}
}

\newcommand{\assignreview}[1]{%
    \ifthenelse{\boolean{showannotations}}%
    {\textbf{\noindent\ding{46}\textcolor{white}{~\colorbox{\assignementcolor}{To review:}}~\textcolor{\assignementcolor}{#1}\\}%
    }
    {}
}

\newcommand{\rem}[1]{%
    \ifthenelse{\boolean{showannotations}}%
    {\textcolor{\oldtextcolor}{\st{#1}}}%
    {}%
}

\newcommand\add[1]{%
    \ifthenelse{\boolean{showannotations}}%
    {\textcolor{\newtextcolor}{{#1}}}%
    {#1}%
}

\newcommand\addblockbegin{%
    \ifthenelse{\boolean{showannotations}}%
    {\color{\newtextcolor}}%
    {}%
}

\newcommand\addblockend{%
    \ifthenelse{\boolean{showannotations}}%
    {\color{black}}%
    {}%
}

\newcommand\rep[2]{%
    \ifthenelse{\boolean{showannotations}}%
    {\rem{#1}~\add{#2}}%
    {#2}%
}

\newboolean{showannotations}
\setboolean{showannotations}{true} 

\hypersetup{pdfborder={0 0 0}}

\newcommand{\newtextcolor}{blue}
\newcommand{\oldtextcolor}{red}
\newcommand{\assignementcolor}{orange}
\definecolor{highlightcolor}{rgb}{.99, 1, .0}
\sethlcolor{highlightcolor}
\definecolor{orcidlogocol}{HTML}{A6CE39}

\mdfsetup{%
   backgroundcolor=gray!5,
   middlelinewidth=1pt,
   roundcorner=7pt}

\title{Scene Graph-based Driving Scenario Extraction for Automotive Egocentric Datasets}

\author{
Stefan Ramdhan, Kyanna Dagenais, Vera Pantelic, Victor Bandur, Mark Lawford
\\McMaster Centre for Software Certification (McSCert)\\
McMaster University\\
Hamilton, Ontario, Canada\\
\{ramdhans, dagenaik, pantelv, bandurvp, lawford\}@mcmaster.ca
}

\date{August 2026}

\begin{document}

\maketitle
\thispagestyle{plain}
\pagestyle{plain}
\begin{abstract}
Extracting scenarios from unlabelled real-world sensor data streams is a critical but challenging task in the development process of automated driving systems (ADS).
Automatically sifting through large datasets to spatially and temporally locate critical scenarios can enable scenario-based coverage analysis of ADS datasets.
In this paper, we present a method for extracting scenarios from egocentric datasets using scene graphs and Linear Temporal Logic (LTL).
We first process egocentric sensor data and HD maps to generate a sequence of scene graphs representing a driving scenario. 
%
Next, we use LTL to formally specify driving scenarios of interest, then extract all instances of the scenarios from the dataset using an off-the-shelf model checker, which evaluates the LTL formula against the sequence of scene graphs.
Our approach can be used on both simulated and real world datasets.
We evaluate the method on the training and validation datasets from Argoverse 2 consisting of 850 15-second real-world driving logs, and several videos of dashcam footage.
We demonstrate the effectiveness of our approach for extracting and querying scenarios by evaluating against a rule-based benchmark based on track annotations and HD maps.
\end{abstract}
\section{Introduction}

Modern automated driving systems (ADS) contain AI-enabled components that are trained on large scenario datasets representative of ADS' operational design domain (ODD). Furthermore, development of ADS deterministic algorithms often relies on large scenario datasets to, for example, parameterize safety models such as Responsibility-Sensitive Safety~\cite{shalev2017formal}. In the verification and validation (V\&V) phase, it is critical to demonstrate correct, safe, and reliable ADS behaviour in scenarios that are representative of the ODD, including both common and edge case scenarios. 
Therefore, across the development lifecycle of ADS, large volumes of data on scenarios within an ODD are necessary. 
Determining whether a dataset of scenarios sufficiently covers the range of scenarios that can be encountered in an ODD requires extracting details about each of the scenarios, such as the number of instances of the scenario and the conditions in which it occurs. Existing methods do not automatically extract the rich semantics conveyed by video data, such as weather, road conditions, and states of traffic participants that might have a significant impact on the criticality of scenarios (e.g., a school bus' stop sign extended). This information would allow us to extract and analyze important safety critical scenarios, such as a robotaxi passing by a stopped school bus with its stop lights on and the stop sign arm extended. Such behaviour has resulted in the U.S.\ National Transportation Safety Board launching an investigation into Waymo \cite{NTSBHWY26FH007}. The sheer size of these datasets makes it prohibitively expensive to manually interpret data to identify such scenarios. 

Our scenario extraction method uses scene graphs~\cite{scenegraphtutorial2021} as a semantic representation of a dataset. Scene graphs have been actively researched in the ADS field in the past five years to help understand the semantics of driving scenes~\cite{malawade2022spatiotemporal,drayson2023cc,woodlief2025scene}. In particular, 
we build on the work of~\cite{woodlief2025scene} to generate sequences of scene graphs from real-world datasets. The scenarios to be extracted are  formally specified via Linear Temporal Logic (LTL). Although we rely largely on the pipeline of~\cite{woodlief2025scene}, we focus on scenario extraction, not runtime monitoring. We make significant enhancements to the toolchain of \cite{woodlief2025scene} to add support for the extraction of track consistent sequences of scene graphs from real-world video data. We focus on real-world data because it presents a more challenging and pressing problem than extracting from simulation data.

The proposed method extracts scenarios defined at the semantic level from real-world egocentric data. We can determine whether a scenario exists within a dataset and, if so, temporally and spatially localize all instances of it. The method allows for analysis of the extracted scenarios such as  coverage analysis in order to determine how well a scenario is represented in a dataset. The results of the analysis can be used to determine the quality of a dataset and ultimately improve it. The approach is evaluated on 850 15-second real-world driving logs from Argoverse (AV) 2~\cite{Argoverse2} supplemented by several videos of dashcam footage. The results indicate high accuracy when scene graphs are generated from vision data complemented with the state and HD map data. There is room for further improvement as most accuracy issues trace back to the tracking quality and do not present a fundamental problem with our approach. 

The primary contributions of this paper are as follows:
\begin{itemize}
    \item  To the best of our knowledge, our method improves on the state of the art in driving scenario extraction by extracting scenarios conditioned on vision-derived semantics from video data, while also retaining the deterministic, white-box benefits that traditional rule-based methods provide. Specifically, our approach can automatically extract scenarios conditioned on the rich semantics from video data, such as weather, road conditions, and features of objects---something that existing methods do not do.
    \item We significantly improve an open-source scene graph generator (SGG), \textsc{roadscene2vec}~\cite{malawade2022rsv}, to better represent semantics relevant to scenario extraction. In particular, we eliminate two notable limitations of the existing tooling  by (1) enabling object tracking in the sequence of generated scene graphs with track based labeling and (2) refining its coarse lane identification.   
    \item We do not introduce new algorithms---we integrate multiple existing systems with additional scaffolding.
        \item All source code used to generate the results contained in this paper are available as an anonymized github repo\footnote{\url{https://anonymous.4open.science/r/scenario-extraction-8C8E/}} to allow replication and extension of this work.
\end{itemize}

The outline of this paper is as follows. Section~\ref{sec:related} presents relevant background information. Section~\ref{sec:approach} introduces our approach, while Section~\ref{sec:results} evaluates the approach. Section~\ref{sec:discussion} discusses the benefits of the approach, and compares it to other approaches. Finally, Section~\ref{sec:conclusions} concludes with avenues for future work. 
\section{Background and Related Work}\label{sec:related}

\subsection{Scenario Definitions and Modelling}

Standard definitions of \emph{scene} and \emph{scenario} were given by \textcite{ulbrich2015defining}. The primary difference is that a scene is a static snapshot of the semantics of the environment, whereas a scenario is a sequence of scenes over time.


We also adopt the taxonomy of \textcite{menzel2018scenarios} specifying scenarios at three different levels of abstraction -- functional, logical, and concrete. Functional scenarios provide high-level descriptions of operating scenarios\footnote{The term `operating scenario' is used to denote ISO 26262's `operational situation', defined by ISO 26262 as `scenario that can occur during a vehicle's life'. The term  includes not only driving scenarios, but also, for example, parking and maintenance.} via domain entities and their relationships, given in natural language. Logical scenarios describe operating scenarios at the state space level, with precisely defined parameters and their ranges, including their probability distributions, where applicable. As such, they are formally described. Finally, concrete scenarios concretize logical scenarios with concrete parameter values. 




\textcite{neurohr2021criticality} later defined a machine-readable version of functional scenarios, called \emph{abstract scenarios}. While functional scenarios can be defined in controlled natural language, abstract, logical, and concrete scenarios are defined in machine-readable formal languages such as LTL~\cite{woodlief2025scene} or domain-specific languages such as OpenSCENARIO~\cite{openscenario}. 

Abstract scenarios are defined at the semantic level using temporal logic---over entities, relations, and how those relations evolve temporally. State-space variables may be used as needed to describe a scenario formally at the semantic level (e.g., formally defining the longitudinal car-following scenario requires specifying the maximum following distance, which is derived from the state-spaces of the ego and lead vehicle).

\subsection{Scenario Extraction}
Traditional approaches to scenario extraction typically define rules with respect to dataset annotations, or extract features then perform clustering. 
Rule-based methods are among the most common approaches~\cite{nakamura2022defining, woodlief2024S3C,ponn2020identification, karunakaran2022automatic}, as they provide an interpretable way to define scenarios directly in terms of dataset annotations such as tracks, ego vehicle state, or maps. 
However, these methods are limited by the information available in the dataset annotations. In particular, they do not capture valuable semantic information that can be obtained from camera data, such as weather, road conditions, or visual context that is typically absent from track annotations, such as whether a school bus has its stop sign extended.

Machine-learning based approaches typically extract features then cluster those features to group scenarios together~\cite{kheriji2024extracting, weber2023toward}. 
\textcite{kheriji2024extracting} use a Variational Autoencoder to encode time series data into a latent space, then cluster to classify into broad scenario categories. \textcite{weber2023toward} cluster trajectory-annotation grids into scenario clusters in a high dimensional space. 
These approaches can learn useful scenario representations, but some require retraining and additional data to extract scenario classes not seen during training, which can become prohibitively costly. These methods not only suffer from a lack of interpretability, but are also limited by information available in dataset annotations.




Video data contains valuable semantic context that is generally absent from driving annotations, however their high dimensionality makes direct scenario extraction challenging.
More recently, 
Vision Language Models (VLMs) and Large Language Models (LLMs) have been explored for scenario extraction from natural language prompts \cite{davidson2025refav}, but have found that repurposing off-the-shelf VLMs for scenario extraction performs poorly. 
RefProg~\cite{davidson2025refav} synthesizes code to extract scenarios from a natural language prompt using an LLM at extraction time, resulting in non-deterministic code generation executed without verification. We provide a detailed comparison between RefProg's results and ours in Section~\ref{sec:results} and conceptually compare them in Section~\ref{sec:comparison_existing_approaches}.


To the best of the authors' knowledge, no existing scenario extraction method can extract scenarios conditioned on semantics derived from vision, while maintaining deterministic and interpretable extraction logic.


\subsection{Scene Graphs}
A scene graph (SG) is a graph $\mathcal{G} = (\mathcal{N}, \mathcal{R})$, where $\mathcal{N}$ is a set of nodes representing the objects in a given scene, and $\mathcal{R}$ is a set of edges between nodes in $\mathcal{N}$, representing the spatial relationships between the objects in the scene. 

SGs are a sensor-agnostic semantic representation of a scene; therefore, they are a natural structure over which to understand an autonomous system's environment. SGs have been used in scene understanding~\cite{li2017scene}, task and motion planning~\cite{ray2024task}, and subjective risk assessment~\cite{malawade2022spatiotemporal}. To the best of the authors' knowledge, SGs have never been applied to egocentric real-world, vision-based scenario extraction. 

The semantics of a scene are task-dependent~\cite{maggio2024clio}---therefore, there are not many open-source SGGs that are easy to use out of the box for scenario extraction. For scenario extraction in a simulator such as CARLA, one can use Carla Scene Graphs~\cite{woodlief2025carlasgg}, an SGG. 
For real-world egocentric video data, we use RoadScene2Vec~\cite{malawade2022rsv}, which was originally created for learning scene-level embeddings from scene graphs, subjective risk assessment, and collision prediction. 
%
\subsection{Linear Temporal Logic}
Since a sequence of SGs represents the evolution of a situation over time, an appropriate approach to identifying properties of such developments is to evaluate LTL formulae against sequences of SGs~\cite{woodlief2025scene}. LTL~\cite{pnueli1977temporal} is a linear time logic that can be used to formally specify and verify properties of systems, including properties of AVs. It uses atomic propositions and logical operators. The logical operators include \textit{and} ($\land$), \textit{or} ($\lor$), \textit{not} ($\neg$), and the temporal operators include $\mathcal{X}$ (Next), $\mathcal{G}$ (Globally), $\mathcal{U}$ (Until), and $\mathcal{F}$ (Eventually).  The temporal operators of LTL allow for the expression of properties along a sequence of events without an explicit need for a notion of time.  We limit ourselves to LTL on discrete finite traces ($\text{LTL}_f$)~\cite{DeGiacomoVardi2013}.  This lets us reason about scenarios as sequences of scenes with a well-defined start and end. 

\section{Our Approach}\label{sec:approach}

\begin{figure*}[t]
  \centering
  \includegraphics[width=\textwidth]{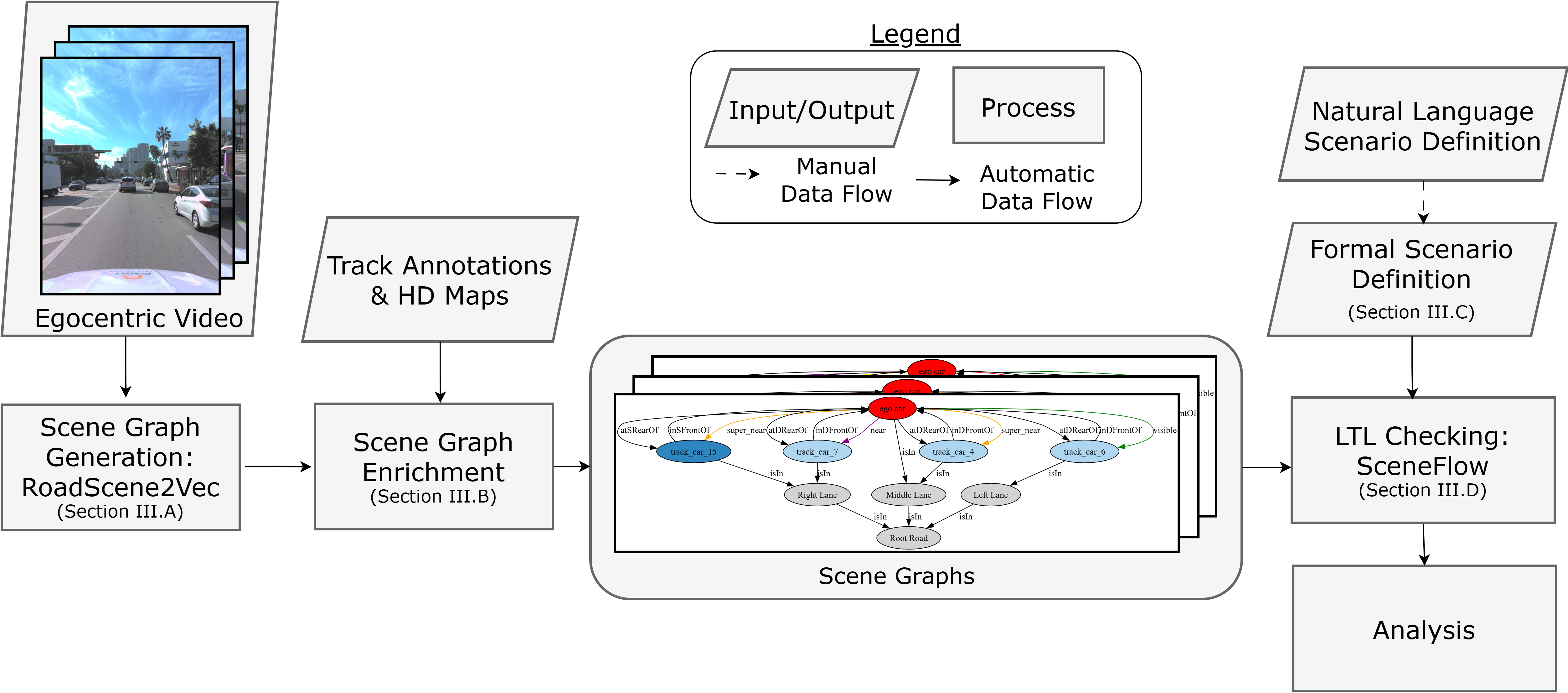} 
  \caption{An overview of the data processing pipeline within our approach.}
  \label{fig:workflow}
\end{figure*}

Our scenario extraction method consists of three steps (see Fig.~\ref{fig:workflow}):
\begin{enumerate}
    \item  We represent a forward-facing camera stream (for extraction from real-world data) as a sequence of scene graphs, where one scene graph represents the relevant semantics of a single frame, i.e., a scene. 
    \item  We formally model a scenario using LTL defined over custom atomic predicates, to capture the desired time evolution of scenes, e.g., a vehicle aggressively cuts in front of the ego vehicle, then brakes.
    \item Then, we evaluate whether the LTL specification is satisfied over the sequence of scene graphs using an off-the-shelf model checker.
\end{enumerate}
  
Our method is able to extract the existence or non-existence of a scenario, its precise beginning and end, and the road user (i.e., track) of interest. The results can be used to enable scenario-based coverage analysis to determine which scenarios are rare or missing in a dataset. An example is given in Table~\ref{tab:scenario_breakdown}. 
The rest of this section describes the scenario extraction process in detail.


\subsection{Scene Graph Generation}

The relevant semantics present in a scene graph depends on the scenarios of interest. For example, a critical scenario may involve a farm animal lying on the road, a school bus stopped with its stop sign extended, or unsecured cargo attached to a moving truck. To reason about scenarios incorporating these elements, a semantic representation must include these semantics in its ontology. For an SGG, this manifests in the ability to express the existence of a farm animal (e.g., as a node in the scene graph), the state of a school bus' stop sign as a node attribute (e.g., schoolBus.stopSignActive), or the relation cargo has to its carrier (e.g., cargo \emph{isSecuredTo} truck). 
A semantically richer scene graph is a means to reason about a scenario at a deeper level, enabling the extraction of a wider range of scenarios.

We use \textsc{RoadScene2Vec (RS2V)}~\cite{malawade2022rsv}, an open-source SGG that can generate a sequence of scene graphs from video data. We use RS2V because it is the only open-source SGG that generates explicit scene graph representations of driving scenes from camera data. 
To generate scene graphs, RS2V first employs Detectron2~\cite{wu2019detectron2}, an object detection and classification model, then a series of rule-based heuristics for determining relationships between nodes. RS2V's ontology is limited to basic spatial predicates (e.g., car\_0 \textit{near} pedestrian\_3,  ego \emph{atDRearOf} (at direct rear of) car\_1, car\_1 \emph{atSFrontOf} (at side front of) ego), and belonging predicates (e.g., ego \textit{isIn} Middle Lane). The full RS2V ontology can be found in~\textcite{malawade2022rsv}. 

Position estimates of objects in the scene graph are generated using a hand-calibrated static birds-eye-view (BEV) projection. The approximate locations of objects' bounding boxes after the BEV projection determine the objects' locations in the BEV projection, which are then used to determine spatial relations and assign entities to lanes. RS2V hard-codes three lanes in all scenarios (labelled ``Left Lane",  ``Middle Lane", and ``Right Lane"), where the ego vehicle is always assigned to the middle lane and any object to the left (right) of the middle lane is labelled as \textit{isIn} the left (right) lane. Not only are the BEV projected object positions noisy, they assume that the ego vehicle pitch is static. When the ego vehicle pitches up or down due to irregularities of the driving surface, the BEV positions become unreliable and thus downstream processing and lane assignments are adversely affected. 

Furthermore, distance relations in RS2V SGs are grouped into boxes  (e.g., ``near'' means dist $\in [18 \text{m}, 30 \text{m})$ and visible means dist $\in [30 \text{m}, 60 \text{m})$). Finally, the spatial relations (e.g., \emph{inDFrontOf}) do not consider object heading---they assume the same heading for all entities. 

As a result of these simplifications, the RS2V SGs are semantically coarse. Scenario extraction via scene graphs generated by vision only is possible, however, in Section~\ref{sec:results} we demonstrate that richer scene graphs are better suited for scenario extraction.


RS2V's ontology is limited to standard elements that are relevant to standard driving scenarios. It serves as a useful starting point to demonstrate scenario extraction for scenarios that involve standard dynamic objects in the automotive context and parameterized edge cases (e.g., lead  vehicle deceleration exceeds 5 m/s$^2$).




\subsection{Scene Graph Enrichment}


Several issues were encountered in attempting to use RS2V to generate sequences of SGs from video data to identify instances of scenarios. Below we describe the issues and how we overcame them. 

\textbf{Object tracking in image space.} Entities within sequences of RS2V SGs generated from video do not contain temporally consistent object labels: an object's label is not preserved from frame to frame. This results in the inability to track an object through a sequence of SGs. 
Additionally, node attributes are limited to estimates of the object's BEV calibrated position and 2D bounding box. 
We implement multi-object tracking using a Kalman Filter and the Hungarian algorithm~\cite{kuhn1955hungarian} to track bounding boxes to generate temporally consistent object labels. We refer to the resulting tracks as perception tracks. 
These temporally consistent SGs are the minimum requirement for extracting scenarios, as scenarios by definition have a temporal element. 

\textbf{State-space \& relation enrichment.} To enrich node attributes with state-space estimates such as position, velocity, and heading, we associate perception tracks to the dataset track annotations containing state estimates. Our method does not require the existence of track annotations---we can employ off-the-shelf tracking and state estimation models to generate state estimates of objects in the ego vehicle's environment, however, we use track annotations for simplicity. 

We project track annotations into image space to obtain a 2D bounding box. We apply the Hungarian algorithm per-frame and per-class (e.g., bus, car, pedestrian) to obtain the annotated track representing the same physical object as the perception track. We allow only one perception track to be associated to an annotated track at any given time. 

Next, the state estimates from the track annotations are added as node attributes. We revise the spatial relations in the graph (e.g., \emph{inSFrontOf}, \emph{near}, etc.) as the state estimates derived from multiple sensors contain lower noise than the relations estimated by RS2V.



\textbf{Map-based Correction. }
We use HD map data if available to remedy the 3-lane simplification employed by RS2V. For each lane and intersection, we add a node that is assigned to a root road. Entities are assigned to lanes using the centroid of their bounding box. We also add graph edges to accurately represent the road geometry in the scene graph (e.g., \textit{opposes}, \textit{toLeftOf}).
If HD map data is not available, lane assignment may be estimated using a vision-based lane detection module.

\textbf{MLLM-based Enrichment. }
%
As mentioned previously, we are interested in extracting scenarios that involve vision-derived semantics that are not present in dataset annotations, such as weather, road surface conditions, and instances of a school bus' stop sign and control arm extended. For this, we use a frontier multi-modal large language model (MLLM) for visual question answering (VQA). Specifically, we use Google's \texttt{gemini-2.5-flash-lite} because it has seen strong performance in weather classification accuracy and schema compliance~\cite{awad2026benchmarking}. We evaluated \texttt{gemini-2.5-flash-lite}'s weather classification capabilities on forward-facing ring camera images from Argoverse 2 using human annotations as a baseline on a statistically significant sample of the dataset, yielding an F1 score of 0.94 and 0.98 on weather and road surface estimation respectively.

For weather and road surface classification, we query the MLLM on a single image, since we use 15-second driving clips, so these conditions are unlikely to change within the short log. To determine whether, in a driving log containing a school bus, its stop sign is extended, the MLLM is queried once per frame, enabling interval extraction. 

Importantly, the MLLM's non-determinism does not carry through to the scenario extraction process. Once the MLLM's answer is used to enrich the scene graph, the scene graphs remain static. This enables our method to retain the determinism and interpretability of the scenario extraction process.


\subsection{Scenario Specification}\label{sec:scenario_spec}
We define a scenario specification as an LTL formula $\phi_{\text{scenario}}$ composed of predicates $\varphi_{1},\dots \varphi_{n}$. These predicates are composed of conjunctions of atomic predicates, which represent simple actions (e.g., ego is moving, vehicle is in right lane). These predicates are defined over scene graph relations, state-spaces, and HD map data depending on the abstraction level of the scenario of interest.
We borrow the convention of~\textcite{woodlief2025scene} to define symbolic entities $e$, which  refer to the same entity across a sequence of scene graphs. 
Our method can extract both abstract and logical scenarios. Recall that abstract scenarios can refer to state-space variables insofar as they are needed to define the scenario semantically.

To illustrate the process of formalizing an abstract scenario in LTL from a functional (i.e., natural language) scenario definition, we use the following example.

\textit{\textbf{Functional Scenario:} Vehicle driving in right lane adjacent to the ego vehicle aggressively cuts into the ego vehicle's lane, resulting in a near collision.}

This scenario can be encoded in LTL by using three distinct events encoded as non-modal predicates: 1) principal other vehicle (POV) driving along in the right lane adjacent to the ego, 2) POV enters ego lane and is partially in ego vehicle's path, and 3) POV is fully in the ego's path and the distance between the two vehicles is less than some threshold. In LTL, the first predicate can be defined as:
\begin{align}\label{eqn:cutin_pred1}
    \varphi_1(d)  \coloneq \text{inAdjacentRightLane}(ego, e) \land \text{sameDirection}(ego, e) \notag\\ \land \text{withinDistance}(ego, e, d) \land \text{isMoving}(ego),
\end{align}
where each term is an atomic predicate defined over the nodes $e \in \mathcal{N}$ and relations $r \in \mathcal{R}$ of the graph $G$ (e.g., $\text{isMoving}(e) \coloneq  \text{abs}(e.\text{velocity}) \geq  \epsilon$), where $\epsilon$ is a buffer to account for noise in the state estimate.
The second and third events can be defined as:
\begin{align}\label{eqn:cutin_pred2}
    \varphi_2 \coloneq \text{vehInEgoLane}(e) \land \text{vehAtSideFrontOf}(ego, e),
\end{align}
\begin{align}\label{eqn:cutin_pred3}
    \varphi_3(d) \coloneq \text{vehInEgoLane}(e) \land \text{vehAtDirectFrontOf}(ego, e) \notag\\ 
    \land \text{withinDistance}(ego, e, d).
\end{align}
where d is a distance threshold: $\text{withinDistance}(ego, e, d) \coloneq (\text{dist}(ego, e) < d)$.

These predicates each represent events that can be checked against a single scene graph. Next, we introduce the temporal operators of LTL to define the whole scenario over predicates $\varphi_1, \varphi_2, \varphi_3$:
%
%

%
\begin{align}\label{eqn:cutin}
\phi_{\text{cut\_in,right\_side}} \coloneq \varphi_1 \land (\varphi_1 \ \mathcal{U} \ (\varphi_2 \land (\varphi_2 \ \mathcal{U} \ (\varphi_3 \land \mathcal{F} \neg \varphi_3)))).
\end{align}
This formula enforces $\varphi_1$ to occur until $\varphi_2$ occurs, then for $\varphi_2$ to persist until $\varphi_3$ occurs, eventually requiring $\varphi_3$ to end, thus concluding the scenario.
Generally, given $n$ total events, one can compose a sequence of events into a scenario specification as,
\[\phi_{\text{scenario}}
\coloneqq \varphi_1 \land (\varphi_1 \ \mathcal{U} \ (\varphi_2 \land (\varphi_2 \ \mathcal{U} \cdots \ (\varphi_n \land \mathcal{F} \neg \varphi_n)))).
\]


Further, to augment the scenario specification with additional attributes such as weather, we define a predicate $\varphi_{\text{weather}}$ to represent the  weather conditions in the scenario, to be evaluated for every SG in a given sequence. For example, we can define a snowy weather predicate, $\varphi_{\text{snowy}}$, as

\begin{align*}
\varphi_{\mathrm{snowy}}(G) =
    \begin{cases}
1, & \text{if } G.\mathrm{weather} = \mathrm{snowy},\\
0, & \text{if } G.\mathrm{weather} \neq \mathrm{snowy}.
\end{cases}
\end{align*}

Then, this predicate is simply conjoined to the existing scenario specification:
\begin{align}\label{eqn:scenario_weather}
    \phi_{\text{scenario,snowy}} \coloneq  \phi_{\text{scenario}} \land \varphi_{\text{snowy}}.
\end{align}
This is the simplest approach, specifying weather that is constant for the duration of the scenario.
In our case, weather information is embedded as metadata in each SG generated, so evaluating the weather predicate is a trivial operation.

Encoding a logical scenario can be done by defining the predicates over ranges of state-space variables. For instance, a scenario defined by the longitudinal deceleration exceeding a threshold $\beta$ can introduce the following predicate:
\begin{align*}
\text{longDecelExceeds}(e, \beta) \coloneq e.\text{longAccel} \leq \beta.
\end{align*}
Then, add a conjunction that becomes true if $\text{longDecelExceeds}$ is true during the scenario $\phi_{\text{scenario}}$:
\begin{align}\label{eqn:lv_decel}
    \phi_{\text{scenario,LV\_decel}} \coloneq \phi_{\text{scenario}} \land \notag\\
    \mathcal{F}(\phi_{\text{scenario}} \land \text{longDecelExceeds}(e, \beta)).
\end{align}

\subsection{LTL Checking}\label{sec:model_checking}

The problem of checking LTL formulae over sequences of SG is a specific, simple instance of the model checking problem.  To evaluate whether an LTL scenario specification is satisfied by a sequence of SGs, we use SceneFlow~\cite{woodlief2025scene}, an implementation of a deterministic finite automaton that flags violations of a specification. We modify SceneFlow to instead flag specification satisfactions as opposed to violations. For more detail on SceneFlow, we refer the reader to~\textcite{woodlief2025scene}.
\section{Evaluation \& Results}\label{sec:results}

\begin{figure*}[t]
  \centering

  \begin{subfigure}[t]{0.49\textwidth}
    \centering
    \includegraphics[width=\linewidth]{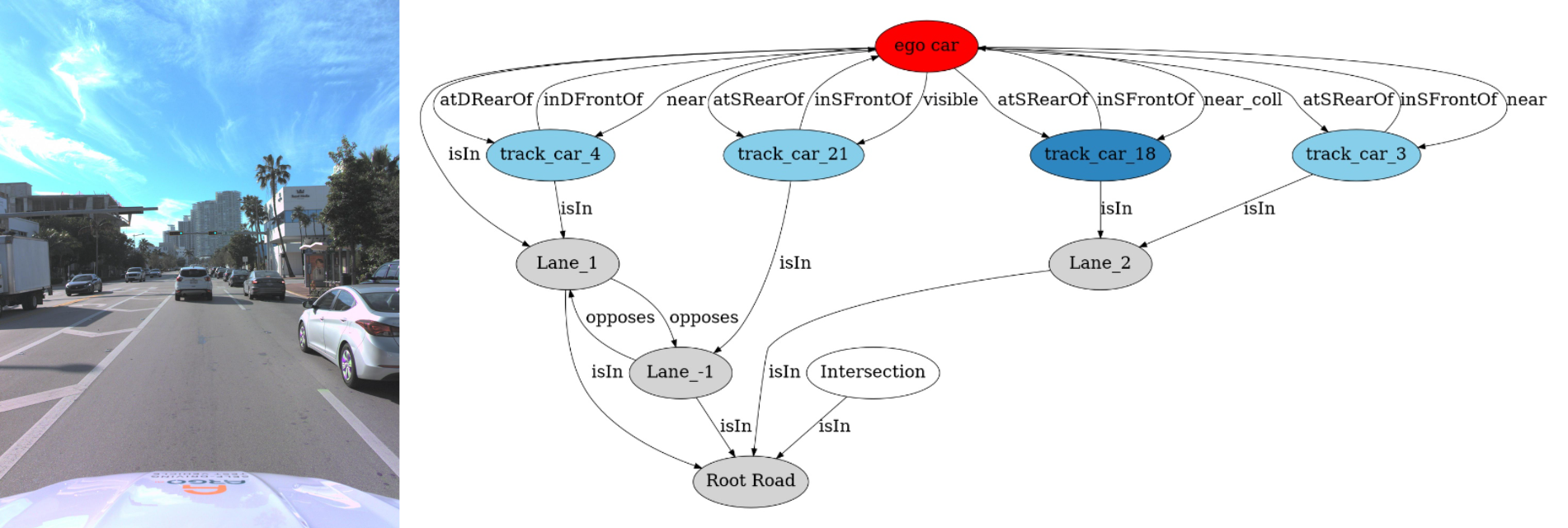}
    \caption{Keyframe when $\varphi_1$ becomes true.}
    \label{fig:img1}
  \end{subfigure}\hfill
  \begin{subfigure}[t]{0.49\textwidth}
    \centering
    \includegraphics[width=\linewidth]{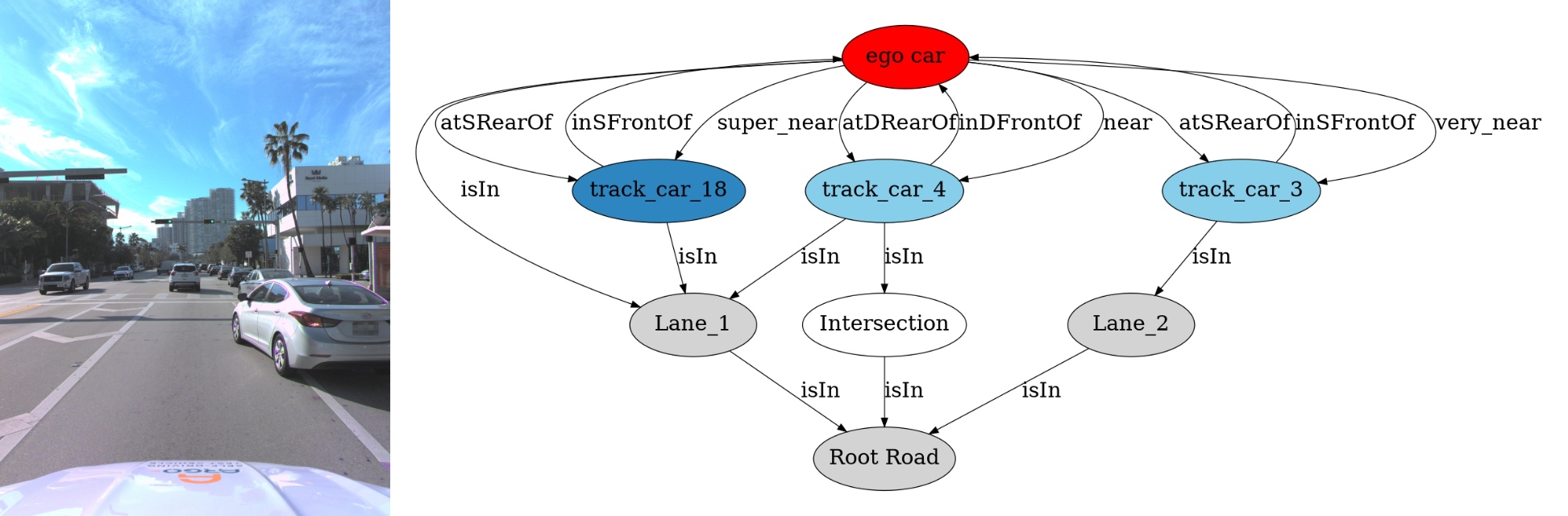}
    \caption{Keyframe when $\varphi_2$ becomes true.}
    \label{fig:img2}
  \end{subfigure}

  \vspace{0.6em}

  \begin{subfigure}[t]{0.49\textwidth}
    \centering
    \includegraphics[width=\linewidth]{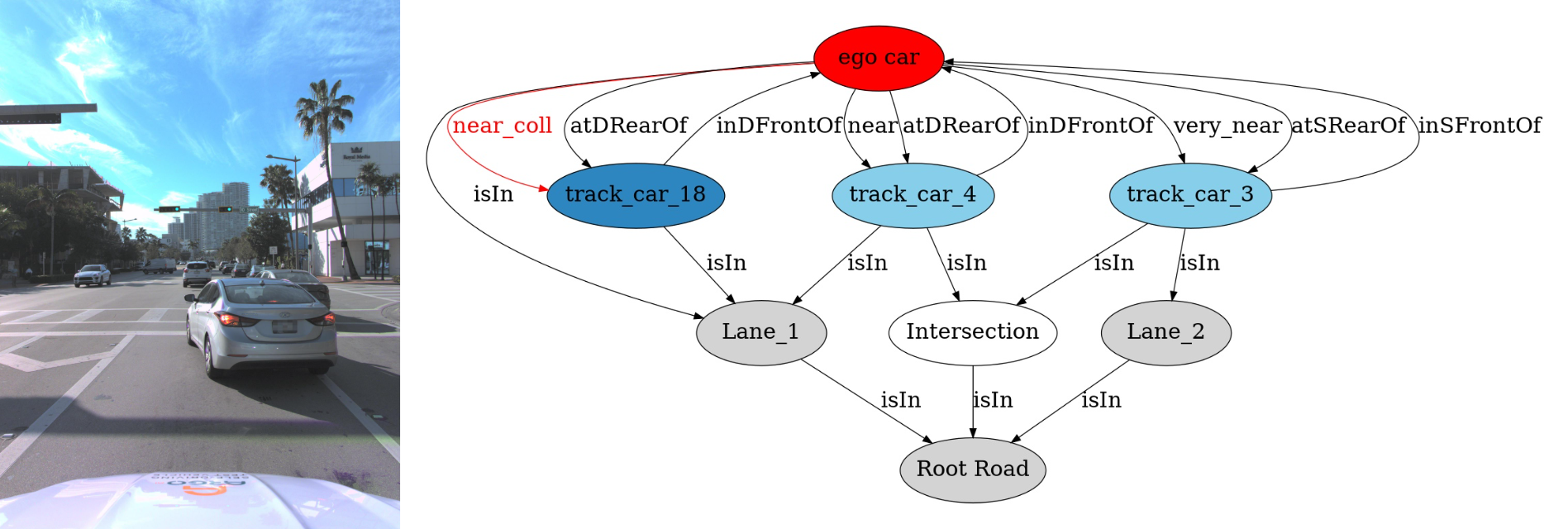}
    \caption{Keyframe when $\varphi_3$ becomes true.}
    \label{fig:img3}
  \end{subfigure}\hfill
  \begin{subfigure}[t]{0.49\textwidth}
    \centering
    \includegraphics[width=\linewidth]{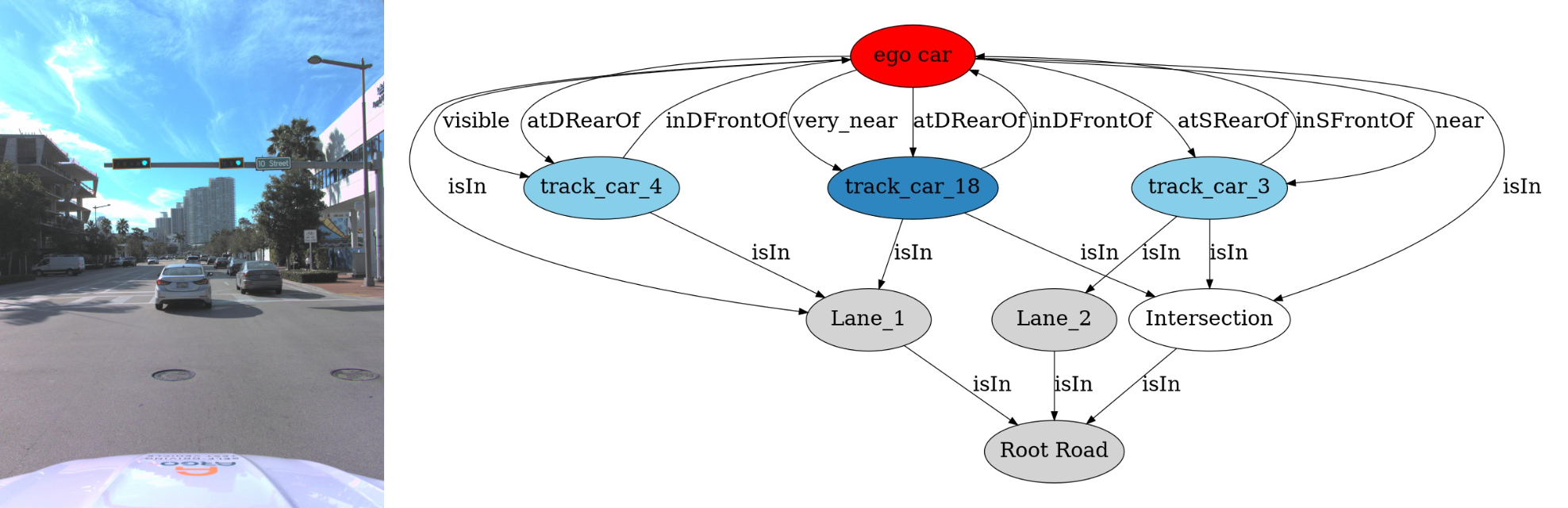}
    \caption{Keyframe when $\varphi_3$ becomes false, upon which the scenario specification is satisfied and our pipeline flags the scenario as having occurred.}
    \label{fig:img4}
  \end{subfigure}

  \caption{Keyframes from an aggressive cut-in scenario, $\phi_{\text{cut\_in}}$ with corresponding scene graphs underneath. The scene graphs are heavily pruned for readability. Given predicates $\varphi_1$, $\varphi_2$, and $\varphi_3$ defined in Equations~\ref{eqn:cutin_pred1}--~\ref{eqn:cutin_pred3}, respectively, $\phi_{\text{cut\_in}}$ is characterized by the activation of the predicates. }
  \label{fig:four-panel}
\end{figure*}

To demonstrate and evaluate the approach, we use the training and validation splits of the Argoverse (AV) 2~\cite{wilson2023argoverse} dataset because of its emphasis on critical driving scenarios. These driving logs are 15-seconds long and include track annotations and HD maps, which are used in the pipeline to enrich the scene graphs.

Though some AV 2 driving logs contain school buses, none of the school buses stop and extend their stop sign. Therefore, we collected several videos of dashcam footage from online sources containing a school bus present with its stop sign extended. We applied two criteria to filter the videos: i) the video must be steady---videos filmed on a mobile device are not applicable and ii) the footage must clearly depict a school bus stopped with its stop sign extended, legible to a human viewer. This process yielded seven dashcam videos. 
For these clips, sensor and map data are unavailable, so scene graphs are produced only from the video stream. To evaluate scenarios involving a stopped school bus, the AV 2 and dashcam footage were combined.


\subsection{Ablation} 

We perform an ablation study to evaluate the relative importance of each step in our pipeline and its impact on scenario extraction performance. We begin with scene graphs that have coarse semantics defined by RS2V's ontology. The only modification made to the raw scene graphs is temporal consistency---the minimum requirement for scenario extraction. We refer to this as Level I in the ablation. Level II scene graphs have been enriched with kinematic data from track annotations. Finally, Level III scene graphs contain kinematic data and HD map data on top of the existing scene graphs.

\subsection{Scenarios of Interest}
\textbf{Longitudinal Car Following: }
To evaluate our method, we extract an abstract longitudinal car following scenario, $\phi_{\text{long,following}}$, defined using a conjunction of predicates,
\begin{align*}
    \varphi_f \coloneq \text{withinDistance}(ego, e, d) 
    \land \text{sameLane}(ego,e) \\
    \land \text{behind}(ego,e) 
    \land  \text{isMoving}(ego).
\end{align*}
We can formally define the scenario as,
\begin{align}\label{eqn:long_following}
    \phi_{\text{long\_following}} \coloneq \varphi_f \ 
    \land \mathcal{F}  \neg \varphi_f.
\end{align}

\textbf{Cut-in from Right Side}: We extract the abstract scenario $\phi_{\text{cut\_in,right\_side}}$ defined by Equation~\ref{eqn:cutin} in Sec~\ref{sec:scenario_spec}. 

In addition, we demonstrate weather extraction by extracting both of the above scenarios under various weather and road conditions. An example is given in Equation~\ref{eqn:scenario_weather}; applying this concept to weather or road conditions is trivial.


We also extract a logical scenario stemming from both of the above scenarios, defined by the POV  decelerating at $\geq 3$ m/s$^2$, which is considered harsh braking. The LTL specification for this scenario is given in Equation~\ref{eqn:lv_decel}, with $\beta = 3$.


\textbf{School Bus Present: }
We follow a similar convention to the longitudinal following scenario, which is defined over a single event. We formally define,
\begin{align*}
    \varphi_p \coloneq \mathrm{schoolBusPresent}(e).
\end{align*}
The scenario is given by,
\begin{align*}
    \phi_{\text{school\_bus\_present}} \coloneq \varphi_p \land \mathcal{F} \neg \varphi_p.
\end{align*}

\textbf{School Bus Stopped: }
Similarly, 
\begin{align*}
    \varphi_s \coloneq \mathrm{schoolBusPresent}(e) \land \mathrm{stopSignExtended}(e).
\end{align*}
The scenario is given by,
\begin{align*}
    \phi_{\text{school\_bus\_present}} \coloneq \varphi_s \land \mathcal{F} \neg \varphi_s.
\end{align*}




\subsection{Scenario Extraction Benchmark}

As AV 2 does not contain ground truth scenario labels, we create a benchmark to evaluate our method. The benchmark is made by defining rules over AV 2 track annotations and HD maps to extract scenarios.
We apply the same definition of long-following programmatically, given in equation~\ref{eqn:long_following} and cut-in from the right side from equation~\ref{eqn:cutin}.

The benchmark for the ``school bus present'' scenario is created by filtering AV 2 tracks for school buses within the forward facing camera's field of view (FOV). Regarding the ``school bus stopped'' scenario, we rely on human annotation to identify the intervals in which the stop sign is extended. 

Verification of scenarios extracted over tracks and HD Maps was verified by a human over a subset of the dataset, as is often done for datasets absent of scenario labels~\cite{davidson2025refav}. For the longitudinal following and cut-in scenarios, 30 scenarios from the benchmark were manually verified. All positive logs for the ``school-bus present'' scenario were manually verified. Since the stopped school bus ground truth was annotated by a human, no manual verification was needed.

The purpose of evaluating our approach against this benchmark is not to demonstrate better performance than a rule-based approach, but to demonstrate agreement with a simple rule-based approach on simple scenarios, while enabling the automatic extraction of scenarios not able to be extracted by the rule-based approach, like weather and road-conditioned scenarios, and the stopped school bus.

The evaluation of our method against this benchmark indicates our method's agreement with a traditional rule-based approach. Since Level I scene graphs are generated over only video data, disagreement (and thus low scores) are expected because they only contain semantics derived from very noisy state estimates, whereas track annotations contain high confidence state estimates derived from multiple sensors. Level II and III scene graphs contain less noisy data, so higher scores are expected.
%
%
In practice, it would only be necessary to extract scenarios using scene graphs generated using a single monocular camera from data with no track annotations nor maps, such as dashcam footage.

\subsection{Metrics}\label{sec:metrics}

\textbf{Tracking Metrics. } Track fragmentation (object's track reported as two or more separate track instances) causes downstream false negatives because scenarios are defined over symbolic entities. When a track ID switches during a scenario, the same physical object is treated as two separate entities, breaking the continuity required for detection. As a result, fragmented tracks can cause valid scenario instances to be missed.  To evaluate track fragmentation and its potential effect on downstream scenario extraction, we measure three quantities: the mean number of perception track fragments associated with each AV 2 track,  the fraction of AV 2 tracks that were represented by multiple perception tracks, and the fraction of an AV 2 track's lifetime is covered by the perception track with the strongest association to it.

Though the tracks in scene graphs have an objectively correct associated AV 2 track, we do not have ground truth associations. We verify whether critical tracks were correctly associated by manual verification, by comparing the temporal evolution of the scene graphs to a visualization of the corresponding scenario over track annotations.

\textbf{Scenario Extraction Metrics.} Spatio-temporal scenario extraction not only identifies the scenario instance, but also the interval in which the scenario is active, and the track of interest in a given scenario. 

We evaluate instance-level extraction using a ``many-to-one" matching approach to account for temporal fragmentation in the extracted interval. We compute the union of all overlapping predicted intervals with the interval of the benchmark. The scenario instance is considered detected if the covered portion of the scenario exceeds a threshold of 50\% and the track of interest has correctly been identified. For example, if the benchmark scenario is active from t=0 to t=10 seconds, but the extracted scenario is active from t=0 to t=4 seconds and from t=6 to t=11 seconds, the union of the overlapping intervals is $\{[0,4], [6,10]\}$. The covered portion of the scenario is 8 seconds, which is 80\% of the scenario interval, thus greater than the 50\% threshold. This extraction is then considered a true positive (TP) extraction if the track of interest is correctly identified. We refer to any metric that requires correctly identifying the track of interest a \textit{track-aware} metric.

We evaluate temporal localization using a track-aware,  timestamp-based interval metric. Rather than treating each extracted scenario interval as a single instance, this metric evaluates agreement at the timestamp level. The union of benchmark intervals defines the positive temporal region, while the union of extracted intervals defines the predicted temporal region. TPs correspond to the total duration where the extracted and benchmark intervals overlap, FPs correspond to extracted duration outside any benchmark interval, and FNs correspond to benchmark intervals not extracted. Precision, recall, and F1 score are then computed from these durations.

For relatively rare scenarios such as cut-ins, most timestamps correspond to negative examples. Inspired by~\textcite{davidson2025refav}, we therefore also report log-balanced accuracy, which averages the true positive and true negative rates over log-level labels, where a log is positive if it contains at least one baseline scenario instance. This metric is non-track-aware and is reported for rare scenarios such as cut-in and school bus stopped; we report here long-following for completeness.

\subsection{Results}
A visual depiction of the cut-in from right side at critical key frames is given in Fig.~\ref{fig:four-panel}. In addition to identifying the start and end timestamps of a scenario, our method extracts key timestamps that correspond to the rising edge of predicates $\varphi_1,\dots,\varphi_n$. 

Tables~\ref{tab:ablation_instance} to~\ref{tab:ablation_log_metrics} show the effect of increasing the semantic richness in the scene graph on scenario extraction. The Level I scene graphs, which are coarse and rely on semantics derived from vision, perform poorly across both instance-level and log-level metrics. This suggests that purely vision-derived scene graphs are too noisy for reliable scenario extraction in their current form. In particular, the log-balanced accuracy for longitudinal following is close to 0.5 in Table~\ref{tab:ablation_log_metrics}. Balanced accuracy averages the true positive (TP) rate and true negative (TN) rate, a score near 0.5 indicates an uninformative classifier. In this case, the low true negative rate suggests that the Level I scene graph frequently predicts longitudinal-following scenarios even in logs where the baseline does not identify one.

A consistent trend across the ablation results is that scenario extraction improves as additional state and map information enrich the scene graph. This is expected, since many driving scenarios depend not only on object identity and coarse semantic class, but also on relative position, lane assignment, heading, and velocity. Adding state information improves the temporal consistency of predicates defined over spatial relations, like relative distance. Adding HD map data improves predicates defined over lane assignments, which are greatly simplified in the Level I scene graphs.

Our scenario extraction approach achieves very strong F1 scores for interval matching and for temporal localization for Level II and III scene graphs. The jump in the long-following F1 score  from 0.552 to 0.827 when ignoring the track-aware constraint (see Table~\ref{tab:ablation_instance}) indicates that, using Level 3 scene graphs, our method extracts the correct interval, but the incorrect track of interest. This is promising, as it indicates that the method is sound, and that many of the false negatives (FNs) and false positives (FPs) stem from issues with tracking quality and association, not as a fundamental limitation of our method.
%

Regarding extraction of $\phi_{\text{school\_bus\_present}}$, both Table~\ref{tab:ablation_instance} and~\ref{tab:ablation_interval} show significantly lower recall than precision (i.e., more FNs than FPs). This is due to the fact that the benchmark identified school buses present as existence within a FOV of the camera, however the camera alone does not reliably detect school busses if they are partially occluded as they often were in many of the logs in which school buses were present. This is simply the result of SGG from a single image; a SGG that works on the inputs from multiple sensors and a more reliable tracking algorithm would not suffer from missing the presence of school buses. 

Temporally extracting instances of school buses stopped is good (see Table~\ref{tab:ablation_interval}, $\phi_{\text{school\_bus\_present}}$'s F1 score of 0.742), however the combination of fragmentation in the detection of the school bus and whether its stop sign is extended compounded into an F1 score of 0.400 in interval matching. In addition to an improved SGG and tracker, we expect the identification of whether the stop sign is extended to improve if multiple frames are fed to the MLLM in sequence. 

Log-balanced accuracy is most relevant for significant class imbalances (i.e., very few or very many instances of a scenario in the dataset). 
In identifying the presence of a school bus, the log-balanced accuracy is near 1 (see Table~\ref{tab:ablation_log_metrics}), though this is mostly downstream from the object detector. Since the scenario is simply defined as the presence of a school bus node in a given scene graph, this perfect score simply indicates that the object detection model did not misclassify the presence of a school bus at the log-level. 

The perfect TP Rate for the school bus stopped scenario indicates that, at the log-level, the MLLM did not miss a single instance of the stop sign extended. However, the TN Rate of 0.992 indicates that the MLLM incorrectly identified a log as containing a school bus with the stop sign extended seven times. Again, querying the MLLM multiple frames at a time or by refining the prompting strategy may improve performance.

For long-following with a 30 and 60 meter following distance and the cut-in scenario, log-balanced accuracy is 0.822, 0.819, and 0.901 respectively when evaluated against Level III scene graphs. 
Although RefProg does not report per-scenario log-balanced accuracy, all three of these values are marginally better than the log-balanced accuracy of RefProg applied to dataset annotations, of 0.811~\cite{davidson2025refav}. 
However, this comparison is not direct, as RefProg's result is aggregated across a different set of complex scenarios rather than reported separately for the three scenarios considered here.

\begin{table*}[h]
\centering
\caption{Ablation study showing many-to-one interval matching  performance using scene graphs with increasingly detailed semantics. A TP extraction must have at least 50\% overlap and identify the relevant track. N indicates the total number of scenarios extracted in the baseline.}
\label{tab:ablation_instance}
\begin{tabular}{llrrrr}
\hline
\textbf{SG Level} & \textbf{Scenario} & \textbf{N} & \textbf{Precision} & \textbf{Recall} & \textbf{F1} \\
\hline

\multirow{3}{*}{Level I: Tracking-only}
& $\phi_{\text{long\_following (30 m)}}$ & 1370 & 0.165 & 0.424 & 0.238 \\
& $\phi_{\text{long\_following (60 m)}}$ & 1948 & 0.193 & 0.393 & 0.259 \\
& $\phi_{\text{cut\_in}}$                & 64 & 0.391 & 0.281 & 0.327\\
& $\phi_{\text{school\_bus\_present}}$                & 54 & 0.969 & 0.574 & 0.721\\
& $\phi_{\text{school\_bus\_stopped}}$                & 7 & 0.308 & 0.571 & 0.400\\
\hline

\multirow{3}{*}{Level II: Tracking \& State Data}
& $\phi_{\text{long\_following (30 m)}}$ & 1370 & 0.453 & 0.397 & 0.423 \\
& $\phi_{\text{long\_following (60 m)}}$ & 1948 & 0.416 & 0.365 & 0.389 \\
& $\phi_{\text{cut\_in}}$                & 64 & 0.847 & 0.781 & 0.813 \\
\hline

\multirow{3}{*}{\makecell[l]{Level III:  Tracking, \\ State, \& Map Data}}
& $\phi_{\text{long\_following (30 m)}}$ & 1370 & 0.632 & 0.583 & 0.607 \\
& $\phi_{\text{long\_following (60 m)}}$ & 1948 & 0.632 & 0.490 & 0.552 \\
& $\phi_{\text{cut\_in}}$                & 64 & 0.943 & 0.781 & 0.855 \\
\hline

\multirow{3}{*}{\makecell[l]{Tracking, State,\\ \& Map  (non-track aware)}}
& $\phi_{\text{long\_following (30 m)}}$ & 1370 & 0.828 & 0.778 & 0.802 \\
& $\phi_{\text{long\_following (60 m)}}$ & 1948 & 0.893 & 0.770 & 0.827 \\
& $\phi_{\text{cut\_in}}$                & 64 & 0.943 & 0.781 & 0.855 \\
\hline


\end{tabular}
\end{table*}

\begin{table*}[h]
\centering
\caption{Ablation study showing timestamp-based temporal localization performance using scene graphs with increasingly detailed semantics. A TP extraction must identify the relevant track.}
\label{tab:ablation_interval}
\begin{tabular}{llrrr}
\hline
\textbf{SG Level} & \textbf{Scenario} & \textbf{Precision} & \textbf{Recall} & \textbf{F1} \\
\hline

\multirow{3}{*}{Level I: Tracking}
& $\phi_{\text{long\_following (30 m)}}$ & 0.394 & 0.781 & 0.523 \\
& $\phi_{\text{long\_following (60 m)}}$ & 0.497 & 0.750 & 0.598 \\
& $\phi_{\text{cut\_in}}$                & 0.459 & 0.315 & 0.373 \\
& $\phi_{\text{school\_bus\_present}}$                & 0.966 & 0.707 & 0.816\\
& $\phi_{\text{school\_bus\_stopped}}$                & 0.924 & 0.620 & 0.742\\
\hline

\multirow{3}{*}{Level II: Tracking \& State}
& $\phi_{\text{long\_following (30 m)}}$ & 0.747 & 0.714 & 0.730 \\
& $\phi_{\text{long\_following (60 m)}}$ & 0.786 & 0.695 & 0.738 \\
& $\phi_{\text{cut\_in}}$                & 0.805 & 0.761 & 0.782 \\
\hline

\multirow{3}{*}{\makecell[l]{Level III: Tracking, State,\\ \& Map}}
& $\phi_{\text{long\_following (30 m)}}$ & 0.788 & 0.789 & 0.788 \\
& $\phi_{\text{long\_following (60 m)}}$ & 0.865 & 0.744 & 0.800 \\
& $\phi_{\text{cut\_in}}$                & 0.941 & 0.755 & 0.838 \\
\hline

\multirow{3}{*}{\makecell[l]{Tracking, State,\\ \& Map  \\(non-track aware)}}
& $\phi_{\text{long\_following (30 m)}}$ & 0.841 & 0.819 & 0.830 \\
& $\phi_{\text{long\_following (60 m)}}$ & 0.931 & 0.779 & 0.848 \\
& $\phi_{\text{cut\_in}}$                & 0.929 & 0.755 & 0.838 \\
\hline

\end{tabular}
\end{table*}
\begin{table*}[t]
\centering
\setlength{\tabcolsep}{3.5pt}
\caption{Ablation study showing log-level scenario extraction performance at different scene graph levels of detail, against the track \& HD Map baseline. N indicates the total number of driving logs containing at least one scenario instance. N$_\text{total}$ is the total number of driving logs used in the evaluation.}
\label{tab:ablation_log_metrics}
\begin{tabular}{llrrrrr}
\hline
\textbf{SG Level} 
& \textbf{Scenario} 
& \textbf{N} 
& \textbf{N$_{\text{total}}$}
& \textbf{Log-bal. Acc} 
& \textbf{TP Rate} 
& \textbf{TN Rate} \\
\hline

\multirow{5}{*}{Level I: Tracking}
& $\phi_{\text{long\_following (30 m)}}$ 
& 555 
& 850 
& 0.539 
& 0.993 
& 0.085 \\
& $\phi_{\text{long\_following (60 m)}}$ 
& 675
& 850 
& 0.514 
& 1.000 
& 0.029 \\
& $\phi_{\text{cut\_in}}$                
& 64 
& 850 
& 0.639 
& 0.311 
& 0.966 \\
& $\phi_{\text{school\_bus\_present}}$                
& 54 
& 857
& 1.000 
& 1.000 
& 1.000 \\
& $\phi_{\text{school\_bus\_stopped}}$                
& 7 
& 857
& 0.996 
& 1.000 
& 0.992 \\
\hline

\multirow{3}{*}{Level II: Tracking \& State}
& $\phi_{\text{long\_following (30 m)}}$ 
& 555 
& 850 
& 0.845 
& 0.706 
& 0.910 \\
& $\phi_{\text{long\_following (60 m)}}$ 
& 675 
& 850 
& 0.813 
& 0.930 
& 0.695 \\
& $\phi_{\text{cut\_in}}$                
& 64 
& 850 
& 0.889 
& 0.787 
& 0.991 \\
\hline

\multirow{3}{*}{\makecell[l]{Level III: Tracking, State,\\ \& Map}}
& $\phi_{\text{long\_following (30 m)}}$ 
& 555 
& 850 
& 0.822 
& 0.913 
& 0.731 \\
& $\phi_{\text{long\_following (60 m)}}$ 
& 675 
& 850 
& 0.819 
& 0.886 
& 0.753 \\
& $\phi_{\text{cut\_in}}$                
& 64 
& 850 
& 0.901 
& 0.803 
& 0.999 \\
\hline
\end{tabular}
\end{table*}

\begin{table}[t]
\centering
\caption{Perception track fragmentation relative to AV2 tracks.}
\label{tab:track_fragmentation}
\begin{tabular}{cccc}
\hline
\textbf{Total \#} & \textbf{Fragments} & \textbf{Fragmented} & \textbf{Dominant Track} \\
\textbf{Tracks} & \textbf{/ Track} & \textbf{Tracks (\%)} & \textbf{Coverage (\%)} \\
\hline
4494 & 1.587 & 34.1 & 63.6 \\
\hline
\end{tabular}
\end{table}

\textbf{Failure Modes.}
Level III scene graphs enable strong performance in scenario extraction, but the results are not perfect. A large source of error comes from the non-deterministic part of the pipeline: object detection and classification, and  downstream processing, like object tracking and association. This can be mitigated by generating scene graphs from multiple images to reduce noise and increase detection confidence.

A main benefit of our approach is that it is white-box and interpretable, making it easy to identify why a false positive or false negative occurred. Misclassifications can be analyzed by inspecting which predicates were active or inactive during the relevant interval, then inspecting the scene graphs and model checking traces to conclude whether the LTL specification was incorrect, or whether the scene graph does not accurately represent the scene.

\textit{False Negatives.} False negatives were commonly caused by track fragmentation. Since scenarios are defined over symbolic entities, a track ID switch during the duration of a scenario causes the same physical actor to be represented as two or more distinct symbolic entities.  As a result, the temporal logic formula may fail even when the underlying driving behaviour is present in the scene. This is particularly problematic for interval-based evaluation, where a fragmented extraction may fail to satisfy the required temporal overlap threshold even if part of the scenario was correctly detected.


\textit{False Positives.} False positives often occurred when the scenario was detected at approximately the correct time, but the wrong actor was identified as the track of interest. These errors are downstream from imperfect track association. 
In such cases, the scenario may be present, but the extracted scenario instance is assigned to the incorrect symbolic entity. This explains why track-agnostic metrics see better results than track-aware metrics. Incorrect track association was especially common in scenes with high traffic density, especially on streets lined with parked cars.

\textit{Detection.} RS2V uses Detectron2~\cite{wu2019detectron2}, an object classification model not trained specifically on driving data. As a result, many instances of one vehicle being detected twice (\textit{e.g.}\ a pickup truck being detected both as a car and a truck), and missed detections---even objects near the ego---were observed. Future work will use an object classification model more suited to the automotive setting.

Regarding MLLM-based VQA, the MLLM was given a single monocular image per query; thus it has no temporal context (e.g., angle of the stop sign changing from previous frames, whether passengers are alighting, etc.)---a key enabler of human VQA abilities especially related to driving tasks. Additionally, in some cases, the school bus was far away from the ego vehicle, or the dashcam video was not of sufficiently high resolution. In some cases, the ``STOP'' text on the stop sign was blurry and the stop sign blended into the background. Additionally, Gemini struggled in instances in which the stop sign is clearly visible but not extended (i.e., school bus heading is orthogonal to ego vehicle heading),  even after revising the prompt to clarify interest in the extension of the stop sign, not the visibility. MLLMs used for VQA in driving scenario understanding may benefit from using video instead of a single monocular image. 



\textit{Tracking. } Tracking issues were mostly downstream of FP and FN detections. We can confirm that tracking is a major source of false negatives both through manual inspection and the tracking quality measurements given in Table~\ref{tab:track_fragmentation}. For each AV 2 track, our tracking system produces an average of 1.587 perception tracks. Moreover, 34.1\% of all AV 2 tracks have more than one perception track associated with it, further indicating that track fragmentation is very common. Finally, the average lifespan coverage of the longest associated perception track is only 63.6\% of the corresponding AV 2 track lifespan, after accounting for the camera field of view. This means that even when the correct actor is detected and associated for part of its visible duration, the dominant scene graph track often does not persist for the full interval over which the actor is relevant. This directly affects scenario extraction, since temporal logic specifications require predicate continuity over time.

\textbf{Coverage Analysis.} One of the motivations of our work is enabling the analysis of scenario-based coverage of egocentric datasets. Table~\ref{tab:scenario_breakdown} demonstrates the extraction of critical scenario variations. By inspecting the model checking traces, we can measure how frequently different scenario variants occur, how long they last, and what fraction of the dataset they cover. Coverage analysis of this sort can provide practitioners with useful analytics for scenario-based validation, enabling the identification of coverage gaps, such as the absence of scenarios in which a vehicle cuts in from the right of the ego in snowy weather, or on icy roads. 

\begin{table}[ht]
\centering
\caption{Breakdown of extracted long-following and cut-in-from-right instances by weather and road condition. Values are reported as \emph{long-following / cut-in from right}. We additionally identify instances in which the lead vehicle (LV) deceleration exceeds $3$ m/s$^2$, including those occurring under adverse road conditions.}
\label{tab:scenario_breakdown}
\begin{tabular}{llr@{\,/\,}rr@{\,/\,}rr@{\,/\,}r@{}}
\hline
\textbf{Scenario}
& \textbf{Breakdown}
& \multicolumn{2}{c}{\makecell{\textbf{\#}\\\textbf{Instances}}}
& \multicolumn{2}{c}{\makecell{\textbf{Median}\\\textbf{Dur. (s)}}}
& \multicolumn{2}{c}{\makecell{\textbf{\% of}\\\textbf{dataset}}} \\
\hline

\textbf{Extracted scenarios}
& \textbf{Total}
& 1304 & 55
& 1.9 & 8.0
& 69.3 & 6.1 \\

\hline

\quad\textbf{Weather condition}
& Sunny
& 872 & 34
& 1.9 & 7.7
& 47.9 & 3.8 \\

& Cloudy
& 422 & 20
& 2.1 & 8.6
& 20.9 & 2.2 \\

& Rain
& 0 & 0
& 0.0 & 0.0
& 0.0 & 0.0 \\

& Snowy
& 0 & 0
& 0.0 & 0.0
& 0.0 & 0.0 \\

& Hazy
& 3 & 0
& 1.4 & 0.0
& 0.1 & 0.0 \\

& Unknown
& 6 & 1
& 5.6 & 12.8
& 0.4 & 0.1 \\

\hline

\quad\textbf{Road condition}
& Dry
& 1190 & 48
& 2.0 & 7.7
& 64.0 & 5.3 \\

& Wet
& 114 & 7
& 1.5 & 10.3
& 5.3 & 0.8 \\

& Snowy
& 0 & 0
& 0.0 & 0.0
& 0.0 & 0.0 \\

& Icy
& 0 & 0
& 0.0 & 0.0
& 0.0 & 0.0 \\

& Unknown
& 0 & 0
& 0.0 & 0.0
& 0.0 & 0.0 \\

\hline

\textbf{LV decel. $> 3.0$ m/s$^2$}
& \textbf{Total}
& 65 & 3
& 4.6 & 12.8
& 5.7 & 0.4 \\

\hline

\quad\textbf{Road condition}
& Snowy
& 0 & 0
& 0.0 & 0.0
& 0.0 & 0.0 \\

& Icy
& 0 & 0
& 0.0 & 0.0
& 0.0 & 0.0 \\

\hline
\end{tabular}
\end{table}

\section{Discussion}\label{sec:discussion}


\subsection{Benefits of the Method}


\textbf{Vision-derived semantics.}
The proposed method is able to extract scenarios from a single monocular camera as demonstrated by the extraction of the stopped school bus. 
The method requires neither HD maps nor track annotations.
However, the larger the volume of available data, the broader the range of scenarios that can be extracted. 
Existing scenario extraction approaches cannot extract the stopped school bus scenario because they are unable to determine whether the school bus' stop sign is extended. This makes our approach very useful in the analysis of the recent set of  incidents involving  Waymo robotaxis passing by school buses without stopping or slowing down.\footnote{\url{https://www.cbsnews.com/news/waymo-investigation-nhtsa-robotaxis-passing-school-bus/}} Additionally, dataset annotations typically do not contain information such as weather, road conditions, or whether a school bus has its stop sign extended; thus, existing scenario extraction approaches are unable to extract scenarios involving these elements. An SGG with an extensive ontology, combined with an MLLM would be capable of producing scene graphs that are a richer semantic representation of a scenario compared to only using dataset annotations. 
%
%

\textbf{Semantic scenario specification.} Scenarios are most naturally reasoned about over semantics---entities, relationships between entities, and how those relations evolve throughout time.
As scene graphs directly contain the semantics of a scene, they are a more natural structure over which to define scenarios. The standardization of relational semantics may enable the reuse of scenario definitions across datasets. Our method can extract abstract scenarios by defining predicates over the scenario semantics present in the scene graph (e.g., relations and how relations evolve temporally), but also logical scenarios by defining predicates over state spaces as traditional rule-based methods do.

\textbf{Determinism. }
The process of scenario extraction is deterministic because it only involves executing the model checker. Given correct scene graphs, LTL specification, and predicate implementations, our method deterministically yields all intervals that satisfy the specification and the track of interest. Determinism is enforced by the intentional separation of non-deterministic processes, such as MLLM-querying, from the extraction itself.

\textbf{Interpretability. }
Our method is white-box and thus interpretable because the model checker outputs detailed traces for every scene graph. This enables troubleshooting errors in the scenario extraction process. Every misclassification can be traced to incorrect i) scene graph(s), ii) predicate implementation, or iii) scenario specification.

\textbf{Spatio-temporal scenario extraction. }
We can query an unlabelled egocentric dataset to determine the extent to which a scenario exists, and details about its coverage to perform a coverage gap analysis. This is absolutely critical in the validation process of an ADS, where automotive companies often have petabytes of egocentric data and want to determine the coverage of a particular set of scenarios.
%
Manually sifting through millions of hours of data is untenable. 
%

\textbf{One-time SGG cost. } The proposed method does not generate scene graphs at extraction time---scene graphs only need to be generated once and are stored as a low dimensional, semantic representation of the dataset. Then, the model checker evaluating the formal scenario definition against the stored scene graphs is fast on a modern personal computer.

\subsection{Relationship to Existing Approaches}\label{sec:comparison_existing_approaches}

\textbf{Vision-derived semantics. } To the best of our knowledge, existing methods are unable to extract scenarios conditioned on vision-derived semantics while retaining the determinism and interpretability of extraction. To extract scenarios conditioned on weather, RefProg~\cite{davidson2025refav} requires the manual annotation of weather in AV 2 driving logs, whereas our approach does MLLM-querying and scene graph enrichment automatically. 

\textbf{Semantic scenario specification. } Rule-based methods and feature clustering methods do not support the specification of scenarios at the semantic level. Scene graphs are an abstraction layer over sensor inputs; thus, they can be a standardized data structure over which scenarios are defined more naturally compared to writing bespoke, hard-coded scripts over tracks and HD maps. RefProg and the submissions to the Scenario Mining Challenge~\cite{davidson2025refav} support semantic scenario specification in natural language, whereas our approach supports semantic scenario specification in LTL. 

\textbf{Determinism. } RefProg uses an LLM to synthesize code from a natural language prompt; thus the same prompt can yield different code, which may extract different scenario intervals and tracks of interest. RefProg code is also executed without verification, so if the synthesized program contains syntactic or semantic errors, the user must either re-prompt the LLM or fix the code manually.

\textbf{Interpretability. } Rule-based methods, RefProg, and the proposed method allow the user to verify the scenario implementation in code. Additionally, each misclassification in the extraction process can be tracked to the input that caused it. 
Neural network-based approaches that perform feature clustering are difficult to troubleshoot because clustering often happens in high dimensional space~\cite{kheriji2024extracting, weber2023toward}.

\textbf{Temporal Scenario Specification. }
RefProg supports the synthesis of a formal scenario specification using boolean logic (e.g., and, or, not) and atomic predicates, whereas our approach uses LTL and atomic predicates. LTL enables specifying the precise temporal scenario evolution, whereas using only boolean logic one cannot specify temporally ordered events.

\textbf{Reusability of scenario definitions. } Rule-based methods tend to rely on bespoke scripts that define scenarios directly over tracks and HD maps, so scenario definitions may have to be changed if a new dataset is used, especially if the scenario is defined with respect to variables that are not present in every dataset. RefProg defines atomic predicates, which were shown to be applicable to datasets other than Argoverse 2~\cite{davidson2025refav}. Our approach relies on atomic predicates that are defined over scene graph attributes. Using a standardized scene graph representation, the same predicates can be reused across datasets. Both methods require a dataset abstraction layer, but do not require redefining scenarios per dataset.

\textbf{Marginal effort of scenario specification. } The process of defining a scenario in LTL is labour intensive, often requiring many iterations and an existing scenario for the purpose of iteratively refining a scenario definition. RefProg synthesizes a formal scenario definition from natural language using an LLM, so the marginal effort required to specify a new scenario is low. However, our approach can serve as a natural bridge to incorporate the use of LLMs for LTL and code synthesis, while retaining the benefit of determinism and interpretability of the extraction inherent in our approach. 

\subsection{Considerations of SGG and LTL}

The diversity of scenarios that can be extracted using our method depends on the ontology of the SGG. The object detection model employed by the SGG must detect scene elements that are relevant to the scenario of interest, and the space of possible relationships between entities must also be relevant to the scenario of interest. For example, if the SGG cannot detect that a stop sign is directly attached to a school bus, or that the school bus has its stop sign activated, extracting the school bus stopped scenario is not possible.
Moreover, the process of formalizing a scenario in LTL requires expert knowledge, can be time consuming for complex scenarios, and is error-prone. However, once a scenario has been formalized in LTL, it can be reused. We envision a library of LTL patterns for simple actions (i.e., atomic predicates), similar to what has been proposed for general LTL properties~\cite{dwyer1999patterns}.

\subsection{Threats to Validity}

Our evaluation dataset is midsize and collected in North America only; however, it contains diverse scenarios, collected across different seasons, weather conditions, and times of day. It also contains urban scenarios as the most challenging to extract and analyze due to traffic density and complex road topologies. 
We plan to expand the set of  scenarios to be extracted and build a library of LTL-specified scenarios. 

While the approach is highly sensitive to the quality of SGG, the results can only improve as the SGG tooling is improved. 

Additionally, the absence of ground truth scenario labels required reliance on a rule-based benchmark constructed for this paper. Though not ground truth, manual verification of numerous scenarios increased confidence in the validity of the benchmark.

\section{Conclusions}\label{sec:conclusions}

We demonstrated the efficacy of temporal logic queries over sequences of scene graphs for extracting scenarios from egocentric video data. The results indicate that our proposed method of scenario extraction from real world video data logs and HD maps is a viable option for the identification of scenario instances and can enable coverage analysis at the scenario level of ADS datasets, including considering weather and road conditions. The efficacy of scenario extraction is heavily dependent on the ontology employed by the SGG, and the semantics extracted from video data. The reasoning nature of this task leads us to believe that VLMs can be utilized to generate scene graphs, or improve the extraction of relevant semantics for existing scene graphs~\cite{toledo2025monitoring}. Additionally, LLMs may be useful in the construction of predicates and LTL specifications from the natural language. This will be considered in future work. Future work will also include scenario extraction of scenarios defined using a more expressive formal language such as OpenSCENARIO, and approaching the problem using the traditional model checker SPIN due to its long and successful track record in verifying safety-critical systems, in order to improve assurance of scenario extraction for high-criticality applications.

\printbibliography
\end{document}